\documentclass[twocolumn]{article}

\usepackage[letterpaper,top=2.5cm,bottom=2cm,left=2.5cm,right=2cm,marginparwidth=1.75cm]{geometry}

\usepackage{indentfirst} %首行缩进

\usepackage{abstract} %摘要格式包

\usepackage{graphicx} %插入图片
\usepackage{amsmath} %公式换行

\usepackage{multicol} %双栏
\usepackage{booktabs} %控制表格线的粗细
\usepackage{color} %字体颜色

\usepackage{natbib} %引用
\setcitestyle{round}
\usepackage[backref]{hyperref}

\usepackage[utf8]{inputenc}

\title{\textbf{Sentence Simplification via Large Language Models}}
\usepackage{authblk}
\author[1]{Yutao Feng}
\author[1]{Jipeng Qiang}
\author[1]{Yun Li}
\author[1]{Yunhao Yuan}
\author[1]{Yi Zhu}

\affil[1]{\textbf{College of Information Engineering, Yangzhou University}}
\affil[]{fyt1600513459@yeah.net, \{jpqiang, liyun, yhyuan, zhuyi\}@yzu.edu.cn}
\date{} %去掉默认时间

\begin{document}

\maketitle

\begin{abstract}
\normalsize
Sentence Simplification aims to rephrase complex sentences into simpler sentences while retaining original meaning. Large Language models (LLMs) have demonstrated the ability to perform a variety of natural language processing tasks. However, it is not yet known whether LLMs can be served as a high-quality sentence simplification system. In this work, we empirically analyze the zero-/few-shot learning ability of LLMs by evaluating them on a number of benchmark test sets. 
Experimental results show LLMs outperform state-of-the-art sentence simplification methods, and are judged to be on a par with human annotators. 
\end{abstract}

\section{Introduction}

Sentence Simplification (SS) is a task of rephrasing a sentence into a new form that is easier to read and understand while retaining its meaning, which can be used for increasing accessibility for people with dyslexia\citep{rello2013dyswebxia}, autism\citep{evans-etal-2014-evaluation} or low-literacy skills\citep{watanabe2009facilita}.

In recent years, neural SS methods utilize parallel SS datasets to train Sequence-to-Sequence models \citep{Wang_Chen_Rochford_Qiang_2016,zhang-lapata-2017-sentence,zhao-etal-2018-integrating} or fine-tune pretrained language models (e.g. BART\citep{lewis-etal-2020-bart})\citep{martin-etal-2020-controllable,lu-etal-2021-unsupervised-method,martin-etal-2022-muss}. However, much work \citep{woodsend-lapata-2011-learning, xu-etal-2015-problems, Qiang2021Unsupervised} pointed out that the public English SS benchmark (WikiLarge \cite{zhang-lapata-2017-sentence}) which align sentences from English Wikipedia and Simple English Wikipedia are deficient, because they contain a large proportion of inaccurate or inadequate simplifications, which lead to the poor generalization performance of SS methods. 

Large Language Models (LLMs) have demonstrated their ability to solve a range of natural language processing tasks through zero-/few-Shot learning\citep{brown2020language, thoppilan2022lamda, chowdhery2022palm}. Nevertheless, it remains unclear how LLMs perform in SS task compared to current SS methods. To address this gap in research, we undertake a systematic evaluation of the Zero-/Few-Shot learning capability of LLMs, by assessing their performance on existing SS benchmarks. We carry out an empirical comparison of the performance of ChatGPT and the most advanced GPT3.5 model (\textit{text-davinci-003}).

To the best of our knowledge, this is the first study of LLMs's capabilities on SS task, aiming to provide a preliminary evaluation, including simplification prompt, multilingual simplification, and simplification robustness. The key findings and insights are summarized as follows:

(1) GPT3.5 or ChatGPT based on one-shot learning outperform the state-of-the-art SS methods. We found that these models excel at deleting non-essential information and adding new information, while existing supervised SS methods tend to preserve the content without change.

(2) ChatGPT is a monolithic model capable of supporting multiple languages, which makes it a comprehensive multilingual text simplification technique. After evaluating the performance of ChatGPT on the task of simplification across two languages (Portuguese and Spanish), we observed that it surpasses the best baseline methods by a considerable margin. This also confirms that LLMs can be adapted to other languages

(3) By performing human evaluation over LLMs's simplification and human's simplification, LLMs's simplifications are judged to be on a par with human written simplifications.

Additionally, the results of this paper is available by visiting \href{https://github.com/BrettFyt/SS_Via_LLMs}{https://github.com/BrettFyt/SS\_Via\_LLMs}.

\section{Related Work}

\begin{figure*}[t]
\begin{center}
\includegraphics[width=1.0\textwidth]{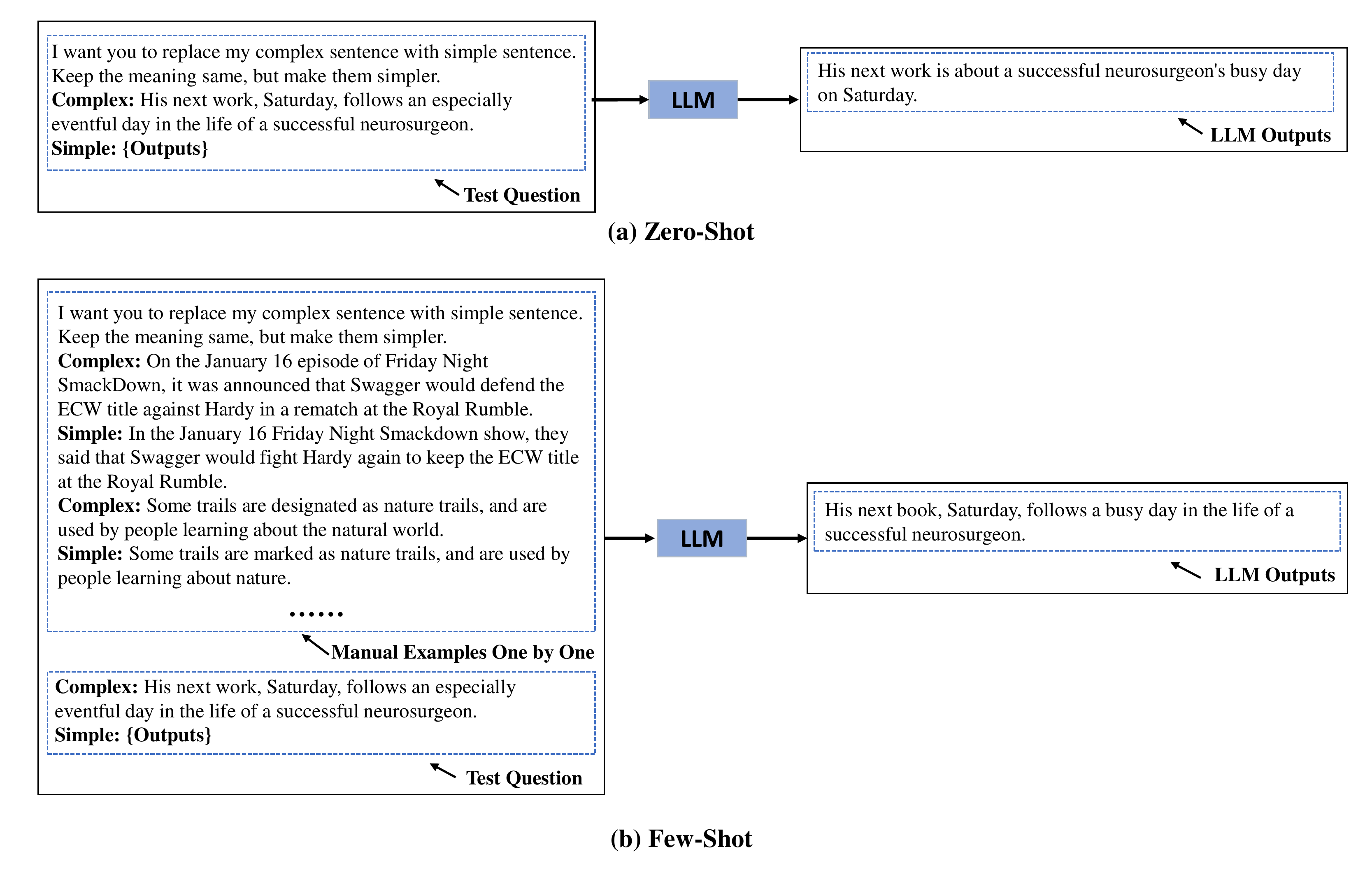}
\end{center}
\caption{(a) is an example of Zero-Shot Sentence Simplification based on LLMs. (b) is an example of few-shot sentence simplification based on LLMs, which stack multiple combinations. \textbf{\{Outputs\}} in the picture means where LLMs output simplified sentences.}
\label{tab:method}
\end{figure*}

\begin{table}[t]
    \small
    \renewcommand\arraystretch{1.2}
    \centering
    \setlength{\belowcaptionskip}{10pt}
    \begin{tabular}{cl}
        \bottomrule[2pt]
        \multicolumn{2}{c}{SS Prompts} \\
        \hline
        ~ & I want you to replace my complex sentence  \\
        ~ & with simple sentence(s). Keep the meaning  \\
        ~ & same, but make them simpler. \\
        ~ & Complex: \textbf{\{Complex Sentence\}} \\
        T1 & Simple: \textbf{\{Simplified Sentence(s)\}} \\
        ~ & ...... \\
        ~ & Complex: \textbf{\{Complex Sentence\}} \\
        ~ & Simple: \textbf{\{Outputs\}} \\
        % \hline
        % ~ & I want you to replace my sentence with the simpler sentence. Keep the meaning same, but \\
        % T2 & make simpler. I want you to only reply the simple sentence and nothing else, do not write \\
        % ~ & explanations. My sentence is ``\textbf{\{Complex Sentence\}}''. \\
        % ~ & \textbf{\{Outputs\}} \\
        \hline
        ~ & Sentence: \textbf{\{Complex Sentence\}} \\
        ~ & Question: Simplify the above sentence \\
        ~ & without changing meaning. \\
        ~ & Answer: \textbf{\{Simplified Sentence(s)\}} \\
        T2 & ...... \\
        ~ & Sentence: \textbf{\{Complex Sentence\}} \\
        ~ & Question: Simplify the above sentence \\
        ~ & without changing meaning. \\
        ~ & Answer: \textbf{\{Outputs\}}\\
        \bottomrule[2pt]
    \end{tabular}
    \caption{Candidate sentence simplification prompts.}
    \label{tab:prompts}
\end{table}

\textbf{Supervised Sentence Simplification methods}:  
Supervised SS methods usually treat SS task as monolingual machine translation, which requires a large parallel corpus of aligned Complex-Simple sentences pairs\citep{nisioi-etal-2017-exploring}. Those methods often employed a Sequence-to-Sequence model (such as Transformer\citep{vaswani2017attention}) as backbone, then integrating different sub-modules into it such as reinforcement learning\citep{zhang-lapata-2017-sentence}, external  simplification rules databases\citep{zhao-etal-2018-integrating}, adding a new loss\citep{nishihara-etal-2019-loss}, or lexical complexity features\citep{martin-etal-2020-controllable}. Another way is to train a sequence editing model, simplifying sentences by predicting the operations for every words\citep{dong-etal-2019-editnts}. These methods rely heavily on public training sets, namely WikiLarge\citep{zhang-lapata-2017-sentence} and WikiSmall\citep{zhu-etal-2010-monolingual}. However, recent studies have pointed out their defects since they contain a substantial number of inaccurate or inadequate simplification pairs \citep{woodsend-lapata-2011-learning, xu-etal-2015-problems, Qiang2021Unsupervised}.

To alleviate this constraint, recent research has concentrated on developing novel parallel SS corpora. In this way, \cite{martin-etal-2022-muss} employed sentence embedding modeling to measure the similarity between sentences from approximately one billion CCNET\citep{wenzek-etal-2020-ccnet} sentences, and subsequently constructed a new parallel SS corpora. \cite{lu-etal-2021-unsupervised-method} utilize machine translation corpus to consturct SS corpora via bach-translation technique. Compared with WikiLarge, these corpora can help to enhance the performance of supervised SS methods.

\textbf{Unsupervised Sentence Simplification methods}: Unsupervised SS methods utilize non-aligned complex-simple pairs corpora. One such method employs style-transfer techniques to achieve content reduction and lexical simplification by importing adversarial and denoising auxiliary losses\citep{surya-etal-2019-unsupervised}. However, this method is less controllable and cannot perform syntactic simplification. Some methods only focus on lexical simplification \citep{qiang2021chinese,qiang2021lsbert}. For pipeline methods, one proposed framework uses revision-based approaches such as lexical simplification, sentence splitting, and phrase deletion\citep{narayan-gardent-2016-unsupervised}, while another improved it by adding an iterative mechanism to these revision-based approaches\citep{kumar-etal-2020-iterative}. However, these methods have too many simplification errors in practice.

\textbf{Large Language Models}: LLMs \citep{brown2020language, thoppilan2022lamda, chowdhery2022palm} have two distinctive features over previous pretrained models. Firstly, LLMs have much larger scale in terms of model parameters and training data. Secondly, unlike previous pretrained models that require finetuning, LLMs can be prompted zero-shot or few-shot to solve a task. Current work\citep{ouyang2022training} shows that the instruct-tuned LLMs are better than finetuned pretrained language models on many natural language tasks. But, there is no work about the capabilities of LLMs on SS task.

\begin{table*}[t]
    \setlength\tabcolsep{4pt}
    \small
    \renewcommand\arraystretch{1.2}
    \centering
    \setlength{\belowcaptionskip}{10pt}
    \begin{tabular}{lcccccccccc}
        \bottomrule[2pt]
        ~ & \multicolumn{5}{c}{TURKCORPUS (En)} & \multicolumn{5}{c}{ASSET (En)} \\
        \hline
        ~ & SARI $\uparrow$ & Add $\uparrow$ & Keep $\uparrow$ & Delete $\uparrow$ & FKGL $\downarrow$ & SARI $\uparrow$ & Add $\uparrow$ & Keep $\uparrow$ & Delete $\uparrow$ & FKGL $\downarrow$ \\
        \hline
        Source & 26.29 & - & - & - & 10.02 & 20.73 & - & - & - & 10.02 \\
        Reference & 40.21 & - & - & - & 8.73 & 45.14 & - & - & - & 6.48 \\
        \hline
        \multicolumn{11}{c}{Supervised Methods} \\
        \hline
        PBMT-R & 38.56 & 5.73 & 73.02 & 36.93 & 8.33 & 35.76 & 4.61 & 59.84 & 42.84 & 8.33 \\
        Dress-LS & 37.27 & 2.81 & 66.77 & 42.22 & 6.62 & 36.90 & 2.40 & 56.14 & 52.15 & 6.62 \\
        % DMASS-DCSS & 39.92 & 7.73 & 38.67 & 7.73 \\
        ACCESS & 41.51 & 6.50 & 71.30 & 46.73 & 7.56 & 40.74 & 6.44 & 62.13 & 53.64 & 7.56 \\
        MUSS-S & \textbf{42.55} & 8.98 & \textbf{73.97} & 44.70 & 7.58 & 44.50 & 11.39 & 62.16 & 59.98 & \textbf{6.40} \\
        \hline
        \multicolumn{11}{c}{Unsupervised Methods} \\
        \hline
        UNTS & 37.20 & 1.50 & 68.81 & 41.27 & 7.84 & 35.19 & - & - & -  & 7.60 \\
        BTTS10 & 36.91 & 2.38 & 69.50 & 38.85 & \textbf{6.30} & 35.15 & 2.21 & 57.91 & 45.32 & 7.83 \\
        MUSS-US & 40.80 & 7.97 & 67.49 & 46.96 & 8.82 & 42.86 & 8.58 & 58.96 & 61.06 & 8.78 \\
        \hline
        GPT3.5 \\
        +Zero-Shot & 37.20 & 8.34 & 49.55 & 53.73 & 6.71 & 44.92 & 10.30 & 53.04 & 71.42 & 6.71 \\
        +Single-Shot & 41.82 & 10.38 & 61.78 & 53.31 & 6.97 & 47.32 & 12.40 & \textbf{61.38} & 68.18 & 6.97 \\
        ChatGPT \\
        +Zero-Shot & 37.72 & 9.89 & 48.82 & \textbf{54.48} & 6.88 & 45.77 & 12.22 & 52.75 & \textbf{72.33} & 6.88 \\
        +Single-Shot & 41.43 & \textbf{11.93} & 58.68 & 54.23 & 7.00 & \textbf{47.94} & \textbf{13.32} & 60.12 & 70.39 & 7.00 \\
        \bottomrule[2pt]
    \end{tabular}
    \caption{Comparison of supervised and unsupervised SS Methods on TURKCORPU and ASSET (En).}
    \label{tab:english_comparison}
\end{table*}

\begin{table*}[t]
    \setlength\tabcolsep{4pt}
    \small
    \renewcommand\arraystretch{1.2}
    \centering
    \setlength{\belowcaptionskip}{10pt}
    \begin{tabular}{lccccccccc}
        \bottomrule[2pt]
        ~ & \multicolumn{4}{c}{ASSET (Pt)} & \multicolumn{5}{c}{SIMPLEXT (Es)} \\
        \hline
        ~ & SARI $\uparrow$ & Add $\uparrow$ & Keep $\uparrow$ & Delete $\uparrow$  & SARI $\uparrow$ & Add $\uparrow$ & Keep $\uparrow$ & Delete $\uparrow$ & FRES $\uparrow$ \\
        \hline
        Source & 20.81 & - & - & - & 5.85 & - & - & - & 51.59  \\
        MUSS-US & 40.94 & 7.71 & \textbf{63.23} & 51.90 & 20.51 & 1.87 & 18.31 & 41.34 & 56.12 \\
        \hline
        ChatGPT & ~ & ~ & ~ & ~ & ~ & ~  & ~ & ~ & ~ \\
        +Zero-Shot & 44.67 & 9.71 & 52.85 & \textbf{71.44} & 38.97 & 6.09 & 26.57 & 84.27 & 66.19 \\
        +Single-Shot & \textbf{47.06} & \textbf{11.30} & 59.20 & 70.71 & \textbf{42.79} & \textbf{7.07} & \textbf{31.40} & \textbf{89.90} & \textbf{70.77} \\
        +Two-Shot & 46.00 & 10.92 & 58.35 & 68.73 & 41.89 & 6.78 & 29.37 & 89.53 & 69.76 \\
        \bottomrule[2pt]
    \end{tabular}
    \caption{Comparison of the state-of-the-art multilingual SS method MUSS\citep{martin-etal-2022-muss} and LLMs SS methods on ASSET (Pt) and SIMPLEXT (Es).}
    \label{tab:other_languages}
\end{table*}

\section{Sentence Simplification via LLMs}

Through exhibiting zero-shot transfer capabilities of LLMs, they have also
become more attractive for lower-resourced tasks. Considering sentence simplification (SS) task lacks large-scale training corpus, we will test the performance of LLMs on SS task.

Specific template patterns, commonly known as prompts, are often employed to guide models towards predicting a particularly desirable output or answer format, without requiring a dedicated training on labeled examples. Utilizing this paradigm shift, we experimented with different prompts issued to OpenAI’s largest available model, GPT3.5 (\textit{text-davinci-003}) and ChatGPT.

\textbf{Simplification Prompts:} To design the prompts for triggering the sentence simplification ability of LLMs, we test multiple prompts to analyze the results. Finally, we meticulously craft two manual instruction prompts, which are illustrated in Table \ref{tab:prompts}.

For the two prompts, \textbf{\{Complex Sentence\}} means the blanks that we need to fill a complex sentence in, while \textbf{\{Outputs\}} was the place carries the outputs of LLMs.

In the first prompt (T1), we utilize the \{Guidance-Complex-Simple\} mapping whereby LLMs are employed to simplify complex sentences into simpler ones under the guidance. In the second prompt (T2), we conceive the \{Sentence-Question-Answer\} mapping methodology to simplify complex sentences in the form of questions. Furthermore, the outputs of the model are unpredictable. In order to achieve a sole output of simplified sentences devoid of any extraneous details, the prompts T1 and T2 employ a specialized guide word, namely ``\textit{Simple:}'' and ``\textit{Answer:}'', respectively, which implement SS by filling in blanks. When executing multilingual SS tasks, we translate the two prompts into the identical languages which utilized in the specific SS tasks.

\textbf{Zero-shot:} When we implement Zero-shot SS, we only need one \{Guidance-Complex-Simple\} combination or one \{Sentence-Question-Answer\} combination for inducing LLMs to generate simplified sentences directly (as shown on Figure \ref{tab:method} (a)).

\textbf{Few-shot:} For few-shot SS, we need to stack multiple combinations for providing simplified examples (as shown on Figure \ref{tab:method} (b)). It is worth noting that we do not need to repeat the guidance in prompt T1, we only need to stack \{Complex-Simple\} combination under the guidance. In contrast, for prompt T1, we need to stack the whole \{Sentence-Question-Answer\}  combination. 

For both prompts, we provide the simplified examples in the part of  \textbf{\{Simplified Sentence(s)\}} for few-shot setting. Since the diversity of simplified forms of a sentence, we also test the way of the multiple manual simplified references. In this way, we change \textit{sentence} in the prompts to \textit{sentences} for adapting to the context of manual references. In actuality, the utilization of multiple simplified references induces the generation of multiple simplified candidates by LLMs. To address this predicament, we opt to select the first simplified candidate for each complex sentence, which yields comparatively superior outcomes in our experiments (Section \ref{Ablation Study}).

\section{Experiments}

\subsection{Evaluation Settings}

\begin{table}[t]
    \small
    \renewcommand\arraystretch{1.2}
    \centering
    \setlength{\belowcaptionskip}{10pt}
    \begin{tabular}{cl}
        \bottomrule[2pt]
        \multicolumn{2}{c}{SS Prompts} \\
        \hline
        ~ & Quiero que reemplaces mis frases complejas \\
        ~ & por frases simples. Mantenga el significado \\
        ~ & sin cambios, pero Hágalo más simple.  \\
        ~ & Complejo: \textbf{\{Complex Sentence\}} \\
        Es & Simple: \textbf{\{Simplified Sentence(s)\}} \\
        ~ & ...... \\
        ~ & Complejo:: \textbf{\{Complex Sentence\}} \\
        ~ & Simple: \textbf{\{Outputs\}} \\
        \hline
        ~ & Quero que substituas a minha frase complexa \\
        ~ & por uma frase simples. Mantenha o mesmo \\
        ~ & significado, mas torne-os mais simples.\\
        ~ & Complexo: \textbf{\{Complex Sentence\}} \\
        Pt & Simples: \textbf{\{Simplified Sentence(s)\}} \\
        ~ & ...... \\
        ~ & Complexo:: \textbf{\{Complex Sentence\}} \\
        ~ & Simples: \textbf{\{Outputs\}} \\
        \bottomrule[2pt]
    \end{tabular}
    \caption{Prompting in Spanish and Portuguese.}
    \label{tab:other_language_prompts}
\end{table}

\textbf{Datasets:} For English (we note as ``En'') SS task, we chose TURKCORPUS (so-called WikiLarge Testset)\citep{xu-etal-2016-optimizing} and ASSET\citep{alva-manchego-etal-2020-asset} as multi-references SS datasets to evaluate the performance of LLMs. Both TURKCORPUS and ASSET are the most popular evaluation datasets for English SS, which consist of 2,000 valid sentences and 359 test sentences. TURKCORPUS employed Amazon Mechanical Turk to give 8 manual simplified versions for each complex sentence. ASSET is an improved version of TURKCORPUS which focuses on the multiple simplification operations, such as lexical paraphrasing, sentence splitting and compression, every complex sentence in it has 10 manual simplified versions sentences.

To evaluate the multilingual generalization efficacy of LLMs, we have opted for the evaluation of multilingual (SS) task in Portuguese (we note as ``Pt'') and Spanish (we note as ``Es''). For Portuguese, we employ  Portuguese version ASSET as our testsets with 359 Portuguese complex sentences, which is also used to evaluate MUSS methods\citep{martin-etal-2022-muss}\footnote{\href{https://github.com/facebookresearch/muss}{This Portuguese dataset can be obtained by visiting the websit: https://github.com/facebookresearch/muss}}. For Spanish, we chose the SIMPLEXT Corpus\citep{saggion2015making} for evaluation. SIMPLEXT is a high-quality 1-to-1 Spanish SS testset with complex 1416 sentences. it is simplified manually by experts for people with learning disabilities. We randomly select 100 sentences from these two datasets for our evaluation and translate prompt T1 into Portuguese and Spanish (as shown on Table \ref{tab:other_language_prompts}).

\textbf{Baseline:} To evaluate English SS task, we compare LLMs with the supervised and unsupervised SS methods. For supervised SS methods, we select three classic methods (PBMT-R\citep{wubben-etal-2012-sentence}, Dress-LS\citep{zhang-lapata-2017-sentence}, DMASS-DCSS\citep{zhao-etal-2018-integrating}, ACCESS\citep{martin-etal-2020-controllable}) and the recent state-out-of-art methods MUSS-S\citep{martin-etal-2022-muss}. For unsupervised SS methods, we compare with three methods (UNTS\citep{surya-etal-2019-unsupervised}, BTTS10\citep{kumar-etal-2020-iterative}, MUSS-Unsup\citep{martin-etal-2022-muss}), where MUSS-US was considered as the state-out-of-art unsupervised SS method. To evaluate multilingual SS task, we also choose MUSS-US\citep{martin-etal-2022-muss}, this method that can complete multilingual SS tasks recently, for comparison.

\textbf{Evaluation Metrics:} To evaluate sentences simplification methods, SARI\citep{xu-etal-2016-optimizing} is primary metric in studies. SARI (the higher the better) compares the generated sentences to the references sentences and returns a arithmetic mean of the n-gram F1 scores of three operations (keeping, adding, and deleting), where $1 \leq n \leq 4$. 

We also report Flesch-Kincaid Grade Level (FKGL)\citep{kincaid1975derivation} to evaluate the readability of the generated sentences. FKGL (the lower the better) is a classic algorithm for measuring readability, which reflects the readability of a text by calculating the age required to understand it. Like the recent works\citep{martin-etal-2020-controllable}, we do not report the BLEU\citep{papineni-etal-2002-bleu} since recent study showed BLEU does not correlate well with sentence simplicity\citep{sulem-etal-2018-bleu}. Since FKGL is not available for Spanish, we report FRES\citep{kincaid1975derivation} instead.

In the experiments, we use standard simplification evaluation package EASSE\footnote{\href{https://github.com/feralvam/easse}{https://github.com/feralvam/easse}} to calculate the SARI and FKGL metrics. We calculate FRES by using the script from \cite{lu-etal-2021-unsupervised-method}\footnote{\href{https://github.com/luxinyu1/Trans-SS/}{https://github.com/luxinyu1/Trans-SS/}}.

\textbf{Other Details:} We chose the latest available LLMs GPT3.5(\textit{text-davinci-003}) and ChatGPT, and we evaluate them by visiting OpenAI’s website\footnote{\href{https://openai.com/api/}{https://openai.com/api/}}, in which calling GPT3.5 requires payment. We set the max length of \textit{text-davinci-003} to 1024 for our few-shot experiments. In addition, for the English few-shot experiments, we randomly select sentences from both TURKCORPUS and ASSET valid sets as simplified examples. Due to the complex-simple pairs in the valid sets of SIMPLEXT is 1-to-1 relationship, so we just use one simplification reference for other languages.

\subsection{Automatic Evaluation} \label{Automatic Evaluation}

\textbf{English Simplification:} Table \ref{tab:english_comparison} shows the results of all SS Methods. As both GPT3.5 and ChatGPT are developed from InstractGPT, they have similar performance in SS tasks. Overall, ChatGPT has demonstrated superior performance to GPT3.5 on ASSET (En) under single-shot, as evidenced by a higher SARI score. This accomplishment has allowed ChatGPT to exceed MUSS-S and establish itself as the new standard of excellence on ASSET (En), achieving an unprecedented state-of-the-art performance increase of +3.44 SARI. Nevertheless, when contrasted with ChatGPT, GPT3.5 and MUSS-S have displayed stronger performance on TURKCORPUS. 

After scrutinizing the trial scores of the simplification operations, we have deduced that LLMs boast a superior deletion score in comparison to other SS methods. This indicates that LLMs SS methods possess a penchant for excising segments of intricate sentences. So we perceive this phenomenon (LLMs exhibit inferiority to MUSS-S on TURKCORPUS) to have emanated from the circumstance that the simplified sentences in ASSET (En) are less intricate than those in TURKCORPUS. The simplified references of ASSET (En) underwent various simplification operations, whereas TURKCORPUS mostly adhered to the original structure\citep{alva-manchego-etal-2020-asset}. However, if we give a simplified example, LLMs will balance the deleting and keeping of the complex sentence. For addition operations, both GPT3.5 and ChatGPT demonstrate high proficiency in scoring, as they are equipped with vast amounts of knowledge derived from copious datasets.

\begin{table*}[t]
    \renewcommand\arraystretch{1.2}
    \centering
    \setlength{\belowcaptionskip}{10pt}
    \begin{tabular}{ll}
        \bottomrule[2pt]
        Complex Sentence: & The Great Dark Spot is thought to \textbf{represent} a hole in the methane cloud deck \\
        ~ & of Neptune. \\
        GPT3.5: & The Great Dark Spot is \textbf{believed} to \textbf{be} a hole in Neptune's methane cloud deck. \\
        ChatGPT: & The Great Dark Spot is thought to \textbf{be} a hole in Neptune's methane cloud deck. \\
        Reference: & The Great Dark Spot is thought to \textbf{be} a hole in the methane cloud deck of Neptune. \\
        \toprule[0.5pt]
        Complex Sentence: & Their eyes are \textbf{quite small}, and their \textbf{visual acuity} is poor. \\
        GPT3.5: & Their eyes are \textbf{small} and they have poor \textbf{vision}. \\
        ChatGPT: & Their eyes are \textbf{small} and their \textbf{vision} is not very good. \\
        Reference: & Their eyes are \textbf{small} and their \textbf{vision} is poor. \\
        \toprule[0.5pt]
        Complex Sentence: & He \textbf{settled in} London, \textbf{devoting himself chiefly to practical teaching}. \\
        GPT3.5: & He \textbf{moved to} London and \textbf{focused mainly on teaching}. \\
        ChatGPT: & He \textbf{lived in} London and \textbf{focused on teaching}. \\
        Reference: & He \textbf{moved to} London and \textbf{focused on teaching}. \\
        \toprule[0.5pt]
        Complex Sentence: & Ned and Dan \textbf{advanced to} the police camp, \textbf{ordering} them to surrender. \\
        \textit{GPT3.5}: & Ned and Dan \textbf{went to} the police camp \textbf{and ordered} them to surrender. \\
        ChatGPT: & Ned and Dan \textbf{went to} the police camp \textbf{and told them to give up}. \\
        Reference: & Ned and Dan \textbf{went to} police camp \textbf{and told them to give up}. \\
        \bottomrule[2pt]
    \end{tabular}
    \caption{The examples of single-shot on English SS Task. The words in bold highlight the differences.}
    \label{tab:examples}
\end{table*}

\begin{table}[h]
    \small
    \renewcommand\arraystretch{1.2}
    \centering
    \setlength{\belowcaptionskip}{10pt}
    \begin{tabular}{cccccc}
        \bottomrule[2pt]
        \hline
        Method & A $\uparrow$ & S $\uparrow$ & F $\uparrow$ & Avg. $\uparrow$ & Rank $\downarrow$\\
        \hline
        ChatGPT & 4.10 & \textbf{3.85} & \textbf{4.32} & \textbf{4.09} & 1.92\\
        MUSS-S & \textbf{4.15} & 3.77 & 4.21 & 4.04 & 2.02\\
        Reference & 3.99 & 3.81 & 4.23 & 4.01 & \textbf{1.87}\\
        \bottomrule[2pt]
    \end{tabular}
    \caption{The results of human evaluation on ASSET (En) based on the adequacy (A), simplicity (S), fluency (F), their average score (Avg.) and subjective ranking of referees (Rank).}
    \label{tab:human}
\end{table}

\textbf{Portuguese and Spanish Simplification:} The results are shown on the Table \ref{tab:other_languages}. It is obvious that ChatGPT outperform the SS results of MUSS-US on both Portuguese and Spanish testset. Firstly, owing to the unsupervised nature of MUSS-US, we compare the Zero-Shot approach of ChatGPT with it. ChatGPT's Zero-Shot methodology surpasses MUSS-US significantly in both languages, namely by +3.73 SARI and +18.46 SARI respectively. Moreover, in the realm of single-shot, ChatGPT obtains a SARI score of 46.00 on ASSET (Pt) and 41.89 SARI on SIMPLEXT, surpassing MUSS by a notable 5 and 21 points, respectively. In particular, in Spanish dataset SIMPLEXT, ChatGPT has a huge lead over MUSS-US in both few-shot and zero-shot scenarios. The aforementioned findings indicate that ChatGPT exhibits strong generalization capabilities across various languages, surpassing MUSS in terms of multilingual SS performance.

\begin{table*}[t]
    \setlength\tabcolsep{3.5pt}
    \small
    \renewcommand\arraystretch{1.2}
    \centering
    \setlength{\belowcaptionskip}{10pt}
    \begin{tabular}{lcccccccccc}
        \bottomrule[2pt]
        ~ & \multicolumn{5}{c}{TURKCORPUS (En)} & \multicolumn{5}{c}{ASSET (En)} \\
        \hline
        ~ & SARI $\uparrow$ & Add $\uparrow$ & Keep $\uparrow$ & Delete $\uparrow$ & FKGL $\downarrow$  & SARI $\uparrow$ & Add $\uparrow$ & Keep $\uparrow$ & Delete $\uparrow$ & FKGL $\downarrow$ \\
        \hline
        \multicolumn{11}{c}{GPT3.5} \\
        \hline
        Zero-Shot & 37.20 & 8.33 & 49.54 & 53.72 & 6.71 & 44.92 & 10.30 & 53.04 & 71.42 & 6.71  \\
        Single-Shot\\
        + Single-Ref & 37.78 & 10.36 & 48.93 & 54.04 & 6.82 & 45.68 & 12.27 & 52.99 & 71.72 & 6.82 \\
        + Multi-Refs + No.1 & 41.82 & 10.38 & 61.78 & 53.31 & 6.97 & 47.32 & 12.40 & 61.38 & 68.18 & 6.97 \\
        + Multi-Refs + Random & 41.69 & 10.09 & 61.97 & 53.02 & 7.45 & 46.49 & 11.04 & 61.28 & 67.15 & 7.45\\
        Two-Shot (Single-Ref) & 40.26 & 11.33 & 55.53 & 53.92 & 7.22 & 46.69 & 13.10 & 56.96 & 70.02 & 7.22 \\
        Three-Shot (Single-Ref) & 39.78 & 10.63 & 54.49 & 54.21 & 7.33 & 46.63 & 12.58 & 56.62 & 70.70 & 7.33 \\
        \hline
        \multicolumn{11}{c}{ChatGPT} \\
        \hline
        Zero-Shot & 37.72 & 9.89 & 48.82 & 54.48 & 6.88 & 45.77 & 12.22 & 52.75 & 72.33 & 6.88 \\
        Single-Shot \\
        + Single-Ref & 38.80 & 11.69 & 49.83 & 54.90 & 7.19 & 47.07 & 14.47 & 54.14 & 72.61 & 7.19 \\
        + Multi-Refs + No.1 & 41.43 & 11.93 & 58.68 & 54.23 & 7.00 & 47.94 & 13.32 & 60.12 & 70.39 & 7.00 \\
        + Multi-Refs + Random & 41.13 & 11.28 & 57.79 & 54.32 & 7.08 & 47.28 & 13.02 & 58.46 & 70.38 & 7.07 \\
        \bottomrule[2pt]
    \end{tabular}
    \caption{The ablation experiments on TURKCORPUS (En) and ASSET (En). ``\textit{+ Multi-Refs + No.1}'' means using multiple simplified references and selecting the first simplified candidate as output. ``\textit{+ Multi-Refs + Random}'' means using multiple simplified references too, but selecting the simplified candidate randomly as output.}
    \label{tab:ablation}
\end{table*}

\subsection{Human Evaluation}

\begin{table}[h]
    \setlength\tabcolsep{4pt}
    \small
    \renewcommand\arraystretch{1.2}
    \centering
    \setlength{\belowcaptionskip}{10pt}
    \begin{tabular}{ccccc}
        \bottomrule[2pt]
        ~ & \multicolumn{2}{c}{TURKCORPUS (En)} & \multicolumn{2}{c}{ASSET (En)} \\
        \hline
        ~ & SARI $\uparrow$ & FKGL $\downarrow$ & SARI $\uparrow$ & FKGL $\downarrow$ \\
        \hline
        \multicolumn{5}{c}{Zero-Shot} \\
        \hline
        GPT3.5 + T1 & \textbf{37.20} & \textbf{6.71} & \textbf{44.92} & \textbf{6.71} \\
        % GPT3 w/ T2 & 36.33 & 6.88 & 43.49 & 6.88 \\
        GPT3.5 + T2 & 36.28 & 7.02 & 43.57 & 7.02 \\
        \hline
        \multicolumn{5}{c}{Single-Shot} \\
        \hline
        GPT3.5 + T1 & \textbf{41.82} & \textbf{6.97} & \textbf{47.32} & \textbf{6.97} \\
        GPT3.5 + T2 & 40.65 & 8.29 & 45.00 & 8.29 \\
        \bottomrule[2pt]
    \end{tabular}
    \caption{Comparison of different prompts for GPT3.5 on two datasets.}
    \label{tab:prompts study}
\end{table}

Since automated metrics may be not enough for evaluating sentence generation, we report the results of human evaluation. In the experiment, we only choose the most advanced method MUSS-S and the reference to evaluation.

% \textbf{Evaluation Metrics:} 
Firstly, we followed the evaluation metrics setup in \cite{kumar-etal-2020-iterative} and \cite{dong-etal-2019-editnts}. We measure the adequacy metric (\textit{How many meanings of the original sentence are retained in the simplified sentences?}), simplicity metric (\textit{Is the sentences output by the system simpler than the original sentence?}), fluency metric (\textit{Is the output sentences grammatical or well-formed?}) on five-point scale (1 is the worst, 5 is the best). In addition, we also measured referees' subjective choices (Ranking simplified sentences from No.1 to No.3) to focus on actual usage rather than evaluation criteria.

We select one hundred complex sentences from ASSET (En) randomly, and then randomly arrange the sentences produced by MUSS-S, ChatGPT, and the simplified sentences randomly selected from the reference files. Then we ask three non-native English speakers with medium level to assess the sentences based on these above metrics.

% \textbf{Result:} 
The results are shown in Table \ref{tab:human}. As expressed in Section \ref{Automatic Evaluation}, due to the tendency of ASSET (En) dataset and ChatGPT to expunge superfluous elements in complex sentences, the adequacy metric lags behind MUSS-S in human appraisal. On the contrary, ChatGPT surpasses MUSS-S in the simplicity and fluency metric owing to this attribute. For human preference, the primary option is ChatGPT and reference sentences, as they hold a superior ranking. This implies that ChatGPT aligns better with individuals' proclivity towards simplicity, in contrast to MUSS-S. It all means the SS performance of ChatGPT surpasses that of MUSS-S in terms of human evaluation and is on a par with human written simplifications.

\subsection{Qualitative Study}

To intuitively analyze the sentence simplification capability of LLMs. Tabel \ref{tab:examples} shows some English sentence simplification examples of GPT3.5 and ChatGPT assembled Single-Shot. 

The method based LLMs significantly reduces the linguistic complexity of the complex sentence while retaining its original main meaning. LLMs perform lexical simplification very well, e.g., ``be'' as a simpler substitute for ``represent'', ``vision'' as a simpler substitute for ``visual acuity'', ``went to'' as a simpler substitute for ``advanced to'', etc. For syntactic simplification, LLMs focus on using simpler and more concise syntactic, e.g., simplifying complex structure ``devoting himself chiefly to practical teaching'' to ``focused on teaching'', the form change from the adverbial clause of ``ordering'' to two parallel clauses by using ``ordered'', etc. In summary, we draw the conclusions from these examples that GPT3.5 and ChatGPT can use only one manual SS example to perform both lexical simplification and syntactic simplification, and these simplification operations are similar to the reference sentences.

\subsection{Ablation Study} \label{Ablation Study}

\textbf{Prompts Study:} We compared the performance differences between different prompts (as shown in Table \ref{tab:prompts}) for SS task. Because the SS performance of GPT3.5 and ChatGPT is very close, we choose GPT3.5 as backbone for the prompts experiments. The performance comparison of different Prompts is shown in Table \ref{tab:prompts study}.

Generally speaking, prompt T1 surpasses prompt T2 in terms of efficacy for both Zero-Shot and Single-Shot SS. Specifically, with regard to the SARI metric for TURKCORPUS, T1 exhibits a superiority of approximately 1 point over T2. Similarly, on ASSET, T1 outperforms T2 by approximately 1.5 points and 2.0 points for the Zero-Shot and Single-Shot respectively.

\begin{figure}[t]
\begin{center}
\includegraphics[width=0.45\textwidth]{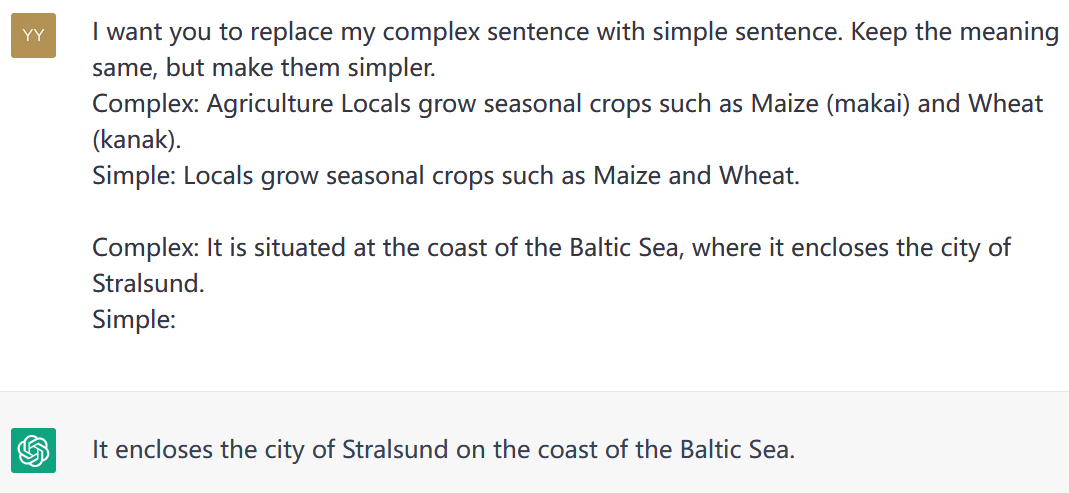}
\end{center}
\caption{The example of single-shot with single-ref.}
\label{tab:single-ref}
\end{figure}

\begin{figure}[t]
\begin{center}
\includegraphics[width=0.45\textwidth]{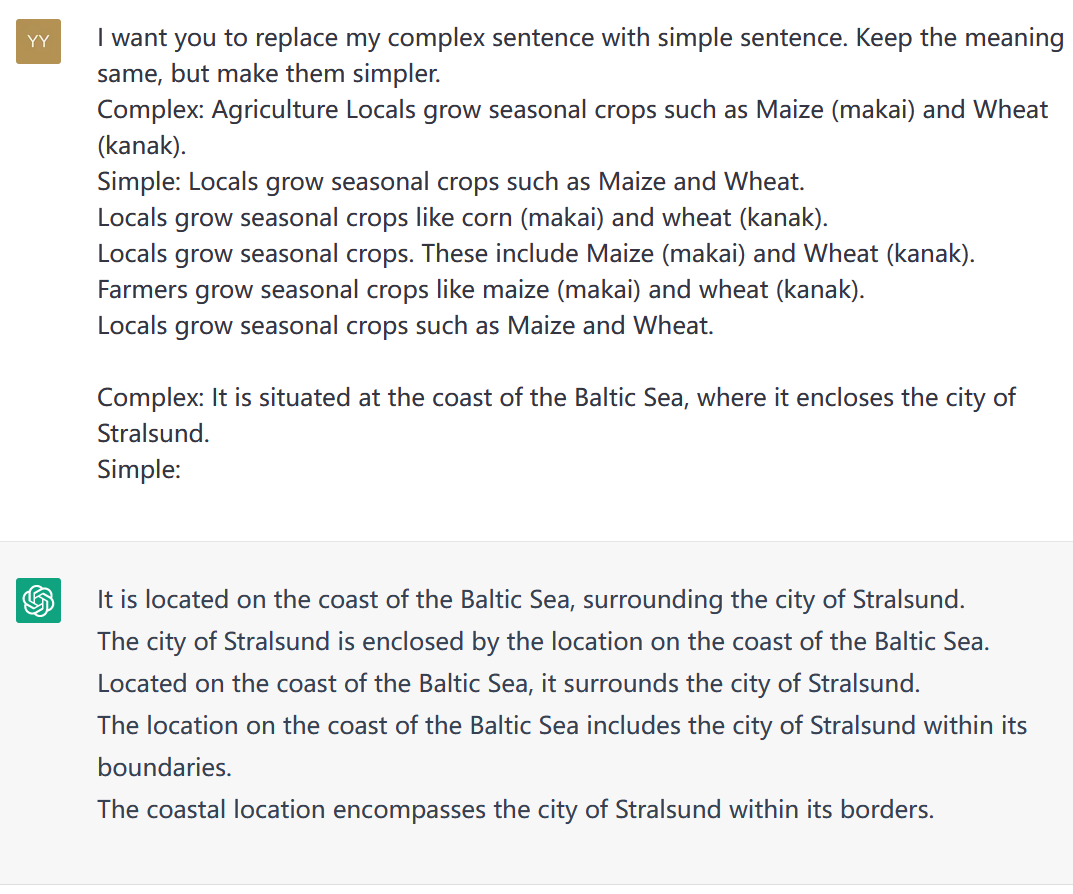}
\end{center}
\caption{The example of single-shot with multi-refs.}
\label{tab:multi-refs}
\end{figure}

\textbf{Few-shot Study:} To further analyze the performance of GPT3.5 and ChatGPT SS methods under mutil-references or Single-reference, we do more experiments in this section. The results of our experiments were shown on Table \ref{tab:ablation}. When juxtaposed with the few-shot method employing a single reference (as shown on Figure \ref{tab:single-ref}), this approach demonstrates substantial improvement in the retention operation of LLMs for intricate sentences, thereby striking a balance between the deletion and retention operations of LLMs (as shown on Figure \ref{tab:multi-refs}). Moreover, in the event that the selection of the first simplified candidate is removed (noted as ``\textit{+ Multi-Refs + Random}''), the performance of SS would diminish, thereby indicating the efficacy of the whole approach, as well as the proclivity of LLMs to furnish the most feasible simplification as a priority in the presence of multiple references.

It is noteworthy that, with an augmentation in the number of Shots, which refers to the increment in the count of sentence simplification examples, there will be a diminishing return in the enhancement of performance. This fact holds true for both the English language and other languages (as shown on Table \ref{tab:other_languages}).

\section{Conclusion}

In this paper, we present a study of the performance of LLMs (GPT3.5 and ChatGPT) for SS task. Given that GPT3.5 and ChatGPT are both derivatives of InstractGPT, their performance in SS tasks is comparable. During the benchmark experiments, LLMs outperformed current state-of-the-art SS methods in the realm of multilingual SS tasks. Furthermore, through the implementation of human and qualitative evaluation, LLMs' simplifications are judged to be on a par with the simplified sentences crafted by human. In our subsequent endeavours, our aim is to design more refined SS methodologies founded on LLMs while also delving deeper into the various proficiencies LLMs offer.

\bibliographystyle{plainnat}
\bibliography{ref}

\end{document}